\documentclass[pdflatex,sn-mathphys-num]{sn-jnl}

\usepackage{graphicx}%
\usepackage{multirow}%
\usepackage{amsmath,amssymb,amsfonts}%
\usepackage{amsthm}%
\usepackage{mathrsfs}%
\usepackage[title]{appendix}%
\usepackage{xcolor}%
\usepackage{textcomp}%
\usepackage{manyfoot}%
\usepackage{booktabs}%
\usepackage{subcaption}
\usepackage{tabularx}
\usepackage{algorithm}%
\usepackage{algorithmicx}%
\usepackage{algpseudocode}%
\usepackage{listings}%
\usepackage{pgfplots}

\theoremstyle{thmstyleone}%

\theoremstyle{thmstyletwo}%

\theoremstyle{thmstylethree}%

\DeclareUnicodeCharacter{009D}{\relax}
\begin{document}

\title[Article Title]{Enhancing Clinician Decision-Making via Uncertainty-Aware Multi-Expert Fusion for Stroke Rehabilitation
}

\author*[1]{\fnm{Tamim} \sur{Ahmed}}\email{tamimahm@usc.edu}

\author[1]{\fnm{Thanassis} \sur{Rikakis}}\email{rikakis@usc.edu}

\affil*[1]{\orgdiv{Bioemdical Engineering}, \orgname{University of Soutehrn California}, \orgaddress{, \city{Los Angeles}, \postcode{90007}, \state{CA}, \country{USA}}}
\maketitle

\begin{abstract}
\noindent
Most stroke survivors live with lasting upper-limb impairment, and tailoring
rehabilitation depends on assessing not merely whether a movement succeeds but
how it is organized. In practice that assessment is the rate-limiting step:
instruments such as the Action Research Arm Test (ARAT) are administered by a
single clinician following standard assessment protocols, compressing a rich
behavioral observation into one ordinal endpoint and discarding the movement-quality
detail that distinguishes recovery from compensation. Automated alternatives have largely chased accuracy on these
noisy single-observer labels and report an opaque point score---the
technology-centric pattern that rarely reaches the clinic. We take a different
route. We present xAARA, an engine that augments rather than replaces clinical
judgment, returning an ARAT assessment from multi-view video with a calibrated
uncertainty and an explanation across task, movement-phase, and movement-quality
levels. Treating scoring as an ill-posed inference under observation uncertainty,
xAARA composes 692 calibrated multimodal models through a Dynamic Bayesian Network
with entropy-based gating, qualifies each result against clinical validity rules,
and defers low-confidence cases. In 88 stroke survivors (788 exercises) it reached
94.2\% task accuracy (Cohen's $\kappa=0.934$) and 81.3\% movement-phase accuracy
($\kappa=0.727$), reducing predictive uncertainty by 96.1\% relative to
conventional single-clinician scoring; where expert scoring was subjective it matched at least
one rater on 100\% of tasks and never returned an out-of-range score. Four
clinicians external to development validated the assessments and were willing to
adopt the system. Principled uncertainty quantification and clinician-aligned
explanation, we argue, are what move automated movement assessment from
demonstration to a deployable clinical tool.
\end{abstract}

\section*{Introduction}

Stroke is a leading cause of long-term disability, and most survivors are left with upper-limb impairment that constrains everyday function~\cite{langhorne2011stroke}. Recovery is greatest in the first weeks but continues for months and differs markedly between individuals, so rehabilitation must be individualized and revised continually as a patient changes~\cite{borschmann2020recovery, li2023stroke}. Doing
that well rests on assessment: the clinician must repeatedly judge not only whether a movement can be performed but also how it is performed, and use that judgment to target and dose therapy~\cite{lang13}. Across much of stroke care, it is this assessment, rather than the delivery of therapy, that is the rate-limiting step.

The judgment that matters most is also the hardest to capture, and this is the core problem our work addresses. A central distinction in upper-limb recovery is between true neurological restitution and compensation — accomplishing a task through altered postures and movement patterns rather than restored control — and the two
carry different prognoses and call for different treatment~\cite{levin2009motor,krakauer2006motor}. Yet the instruments in routine use, such as the ARAT~\cite{mcdonnell2008action,yozbatiran2008standardized} and the Fugl-Meyer Assessment~\cite{gladstone2002fugl}, score whether and how well a task is completed while remaining largely blind to how the movement is organized; they credit a compensatory success as readily as a neurotypical one and are insensitive to the very changes in movement quality that separate recovery from compensation and that mark meaningful progress~\cite{levin2009motor, van2001responsiveness}. Therefore, it is essential to quantify movement quality for measuring sensitive outcomes in individualized care. But it has remained difficult to do reliably and at
scale~\cite{nordin2014assessment, cirstea}.

These limitations are compounded by how the scoring is done. An ordinal score (i.e., in ARAT between 0 and 3) is assigned by a single clinician following standard assessment protocols. This compresses the rich behavioral observation of the experts into one number with limited responsiveness to change and rater-dependent reliability~\cite{van2001responsiveness, lannin2004reliability}---so
much so that standardized administration protocols were introduced specifically to curb the variability between raters~\cite{yozbatiran2008standardized}. This subjectivity is not simply noise to be averaged away: on borderline movements, it reflects genuine ambiguity about which score is correct. Therefore, treating one
rater's label as the single ground truth discards information that a second expert would weigh differently~\cite{dawidskene1979, uma2021disagreement}. Automation has so far sidestepped the
problem. Sensor- and video-based systems built to reproduce ARAT-style scores~\cite{dutta2021poststroke, ahmed2025automatedaratscoringusing} are trained on
a single consensus or majority label and evaluated against a single ground truth. Such pipelines are not capable of resolving the ambiguity; they are an instance of the broader pattern of fitting a model to a single-observer label and reporting accuracy that has produced strong benchmarks yet little clinical impact~\cite{sokol2025translational}. It leaves unaddressed both the explicit quantification of uncertainty~\cite{zhou2025uncertainty} and the explicit handling of the expert subjectivity that a trustworthy assessment requires.

The difficulty is, moreover, intrinsic. Human movement is high-dimensional and redundant, so the same ordinal score can arise from many different movement organizations. Recovering a clinical judgment from video is an ill-posed inverse problem, compounded by occlusion, viewpoint, and genuine ambiguity in 
performance~\cite{TODO:neuromechanics}. Because this expert subjectivity is legitimate rather than an error, a model fit to a single score is chasing a noisy, lossy target. Our response is to
change the objective: not to match or beat the clinician, but to augment the clinician—removing the measurement bottleneck while keeping the clinician as the locus of authority and routing uncertain cases back for review~\cite{sokol2025translational}.
This requires a representation aligned with clinical reasoning, calibrated uncertainty, and an explanation pitched at the level of the decision. We build on a domain-agnostic computational translator that organizes evidence in a nested Environment–Activity–Goal–Meaning (EAGM) representation coupled to a bidirectional Dynamic Bayesian Network~\cite{ahmed2025methodology}.   The translator works at the core of a co-designed multi-view capture-and-annotation system developed with practicing physical therapists~\cite{TODO:xAARA}. The system  converts an essentially unlearnable target into a learnable one by rendering tacit clinical judgment as structured, multi-level annotations and co-designing interventions.

On this foundation, we built xAARA, an engine that assesses each ARAT exercise from multi-view video and returns a task score, a score for each movement phase, the implicated movement-quality elements, and a calibrated uncertainty that determines whether a case is reported or deferred. Its design maps directly onto the rehabilitation problems above. By decomposing performance into movement phases and
quality elements, it surfaces the movement-quality information clinicians use to distinguish compensation from recovery, rather than collapsing it into one number~\cite{levin2009motor, nordin2014assessment}. By composing many calibrated
models and reducing predictive uncertainty by two orders of magnitude, it converts a subjective, rater-dependent score into a quantified one. And by delivering this in minutes per task with clinician-validated agreement, it speaks to the assessment
bottleneck that limits individualized therapy and sensitive trials. In a cohort of 105 stroke survivors, we show that co-designed annotation reduces labeling uncertainty while preserving genuine inter-rater subjectivity. The system's probabilistic composition yields accurate, low-uncertainty, interpretable assessments; that is, in cases where two experts score a movement differently, the family of models reduces this subjectivity by returning a score within the set of expert-endorsed answers rather than an outlier. The engine was judged to be adoptable by clinicians external to its development.

\section*{Results}

\begin{figure}[t]
  \centering
  \includegraphics[width=\textwidth]{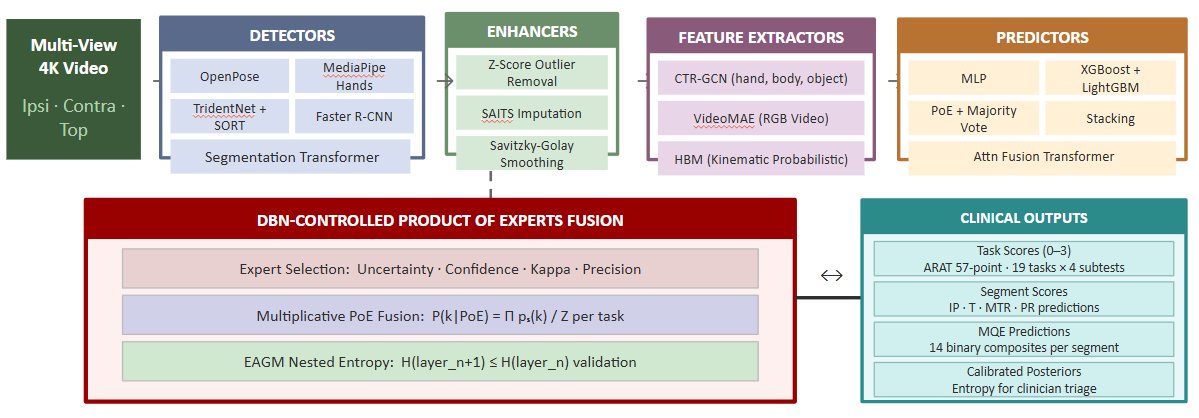}
  \caption{\textbf{The xAARA assessment engine.} Synchronized multi-view video
  (ipsilateral, contralateral-zoom, top-down) flows through detectors, enhancers,
  feature extractors, and predictors; a DBN-controlled Product-of-Experts fusion
  with EAGM nested-entropy validation composes the predictions and emits task,
  movement-phase, and movement-quality outputs with calibrated posteriors for
  clinician triage.}
  \label{fig:pipeline}
\end{figure}

\subsection*{An engine that scores, quantifies its uncertainty, and explains}

xAARA assesses each ARAT exercise from synchronized multi-view video through a five-stage pipeline (Fig.~\ref{fig:pipeline}). Detectors localize body, hand, and object landmarks and segment the exercise into its movement phases; enhancers clean and impute the resulting signals; feature extractors produce skeleton (CTR-GCN), appearance (VideoMAE), and handcrafted-kinematic (HBM) representations;
predictors map these to scores through ten model families, the strongest of which is a Segment-Attention Fusion Transformer that we introduce (SAFT; Methods, Fig.~\ref{fig:saft}); and a DBN-controlled Product-of-Experts fusion composes the predictions with calibrated uncertainty. For each exercise, the engine returns an
assessment spanning all four EAGM layers— a behavioral-context label, a task score, scores for the four movement phases (initial positioning [IP], grasp [T], manipulation-transport [MTR], and placement-release [PR]), and the implicated movement-quality elements—each with a confidence and an entropy that supports clinician triage. In total, the engine is assembled from 692 component models (198 task-level, 62 phase-level, and 432 movement-quality-level) spanning ten
families.

\subsection*{Reducing entropy in the Clinician's Annotations with Interventions}

Because a model can be no better than the labels it is fit to, we first
characterized the uncertainty in the clinician labels themselves. Using a
hierarchical Bayesian label model spanning the task, phase, and movement-quality levels, we tracked how labeling entropy fell as the annotation was enriched (Table~\ref{tab:entropy}). A single therapist scoring under standard protocol had a mean task entropy of 0.459 nats—one third of the maximum for a four-point scale. Recorded multi-view review by two clinicians reduced this to 0.147 nats; adding
explicit movement-phase structure brought it to 0.036 nats; and the full
hierarchical model reached 0.004 nats. Once the evidence clinicians actually use was made expressible, the task score became, for practical purposes, unambiguous.

\begin{table}[h]
  \centering
  \caption{Mean task-level labeling entropy $H(T)$ across annotation stages
  (nats; final column normalized by $\log 4 \approx 1.386$).}
  \label{tab:entropy}
  \begin{tabular}{lcc}
    \toprule
    Annotation stage & $H(T)$ (nats) & Fraction of max. \\
    \midrule
    Single clinician, standard protocol          & 0.459 & 0.33 \\
    Recorded, two clinicians, multi-view         & 0.147 & 0.11 \\
    + movement-phase structure                   & 0.036 & 0.03 \\
    Full hierarchical model                       & 0.004 & 0.003 \\
    \bottomrule
  \end{tabular}
\end{table}

This collapse is not the suppression of expert subjectivity, a distinction that matters both clinically and statistically. Decomposing the residual uncertainty showed near-zero entropy at the task level (0.004 nats) but substantial entropy at the phase (0.093 nats) and movement-quality (0.292 nats, $\sim$42\% of the binary maximum) levels: raters agree \emph{that} an exercise is a 2 while remaining uncertain about \emph{which} quality deficits produced it. Inter-rater agreement was substantial at the task ($\kappa=0.748$) and phase ($\kappa=0.699$) levels and ranged from substantial to fair across the movement-quality composites ($\kappa=0.273$--$0.666$); the composites clinicians agreed on most were also those the models later detected best, locating the difficulty in the movement-quality elements themselves rather than in the annotation or the model. The consequence is direct: holding the pipeline fixed and changing only the labels, training on the co-designed annotations rather than standard single-clinician labels improved accuracy by 8.8 percentage points (87.4\% vs.\ 78.6\%) and reliability far more ($\kappa=0.847$ vs.\ $0.615$). The same model gains most of its performance from a better target, not a better architecture—and supervision at the phase level additionally lifted task prediction (87.4\% vs.\ 84.1\%), the computational echo of the entropy reduction in the human labels.

\subsection*{Composing calibrated models yields accurate, low-uncertainty, interpretable assessments}

With the labels resolved, the engine's accuracy comes from composition rather than from any single estimator. Cross-modal fusion beat every individual modality at each level, and the evidence that mattered tracked clinical reasoning: at the task level, ipsilateral fusion of skeleton and appearance reached 87.4\%
($\kappa=0.847$), above either modality alone (77.5\%, 78.6\%) and a
handcrafted-kinematic baseline (81.6\%); at the phase level, where fine spatial detail dominates, appearance-based fusion led (78.8\%, $\kappa=0.719$); and our SAFT model, which preserves rather than pools the per-phase structure a clinician attends to, was strongest under every condition (Methods). Composed through the full entropy-gated fusion over 88 patients and 788 exercises, the engine reached
94.2\% task accuracy ($\kappa=0.934$, excellent) and 81.3\% phase accuracy
($\kappa=0.727$, substantial; Table~\ref{tab:accuracy}), with residual errors
concentrated at the ordinal 2--3 boundary where near-normal movement is clinically hardest to separate from normal.

\begin{table}[h]
  \centering
  \caption{xAARA performance against clinician consensus.}
  \label{tab:accuracy}
  \begin{tabular}{lcccc}
    \toprule
    Level & $n$ & Accuracy & Cohen's $\kappa$ & Agreement \\
    \midrule
    Task   & 788     & 0.942 & 0.934 & Excellent \\
    Phase  & 2{,}701 & 0.813 & 0.727 & Substantial \\
    \bottomrule
  \end{tabular}
\end{table}

Composition did more than raise accuracy; it made the engine's confidence trustworthy, so that the system reliably knows when it knows. Normalized predictive uncertainty fell from 0.599 for a single model trained on standard single-clinician scores to 0.148 after
multi-view consensus fusion (75.3\%) and to 0.023 under full composition (96.1\%; Table~\ref{tab:uncertainty}). A stage-wise decomposition placed the two largest reductions at the multi-view (Environment) and cross-layer movement-phase (Goal) stages—the same ordering seen in the clinician labels, indicating that the EAGM layers capture the genuine hierarchical organization of the uncertainty rather than an artifact of the model.

\begin{table}[h]
  \centering
  \caption{Normalized predictive uncertainty across rating paradigms
  ($H_{\text{norm}}\in[0,1]$; 88 patients, 3-fold cross-validation).}
  \label{tab:uncertainty}
  \begin{tabular}{lcc}
    \toprule
    Rating paradigm & $H_{\text{norm}}$ (mean$\pm$SD) & Reduction \\
    \midrule
    Single-clinician-trained model  & $0.599\pm0.011$ & --- \\
    Multi-view consensus fusion     & $0.148\pm0.021$ & 75.3\% \\
    Full xAARA composition          & $0.023\pm0.005$ & 96.1\% \\
    \bottomrule
  \end{tabular}
\end{table}

\begin{figure}[t]
  \centering
  \begin{subfigure}[t]{0.97\textwidth}
    \centering
    \includegraphics[width=\textwidth]{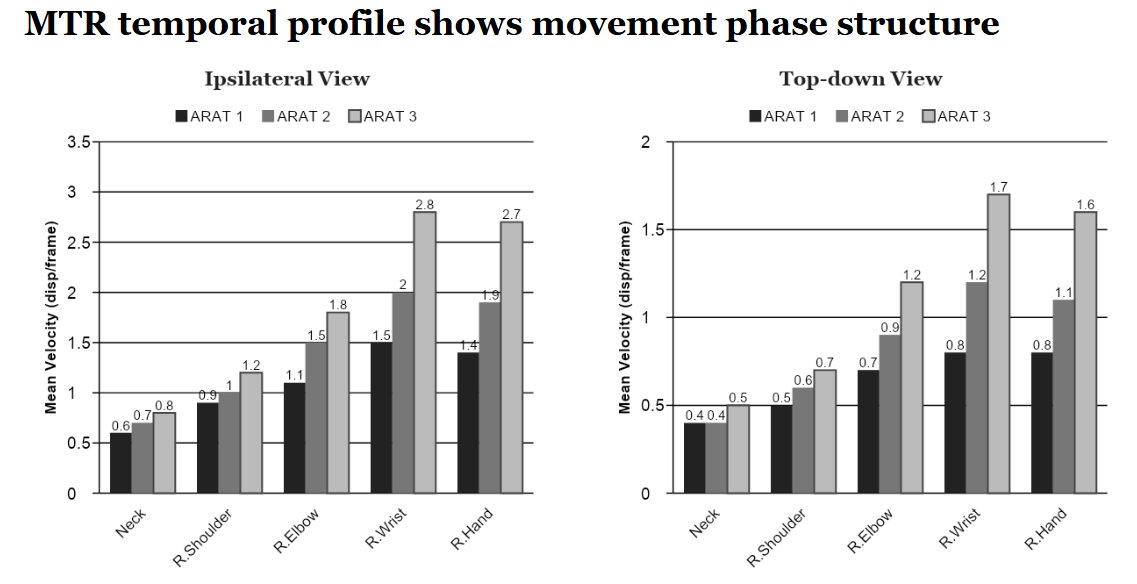}
    \caption{Mean per-joint velocity during manipulation-transport by ARAT score.}
  \end{subfigure}\\[1.5ex]
  \begin{subfigure}[t]{0.49\textwidth}
    \centering
    \includegraphics[width=\textwidth]{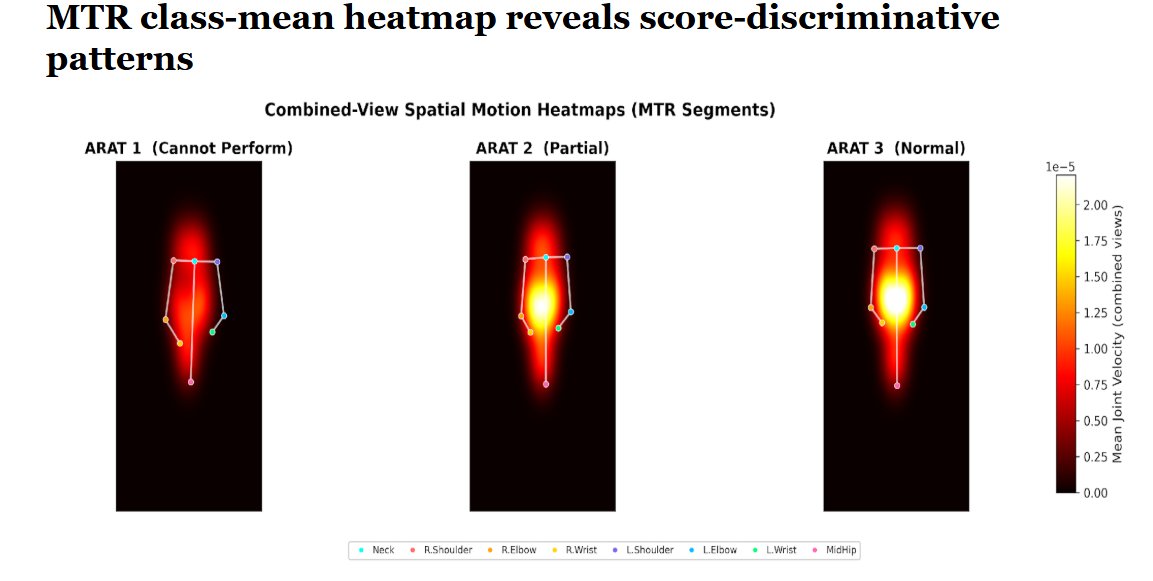}
    \caption{Combined-view class-mean spatial motion heatmaps.}
  \end{subfigure}\hfill
  \begin{subfigure}[t]{0.49\textwidth}
    \centering
    \includegraphics[width=\textwidth]{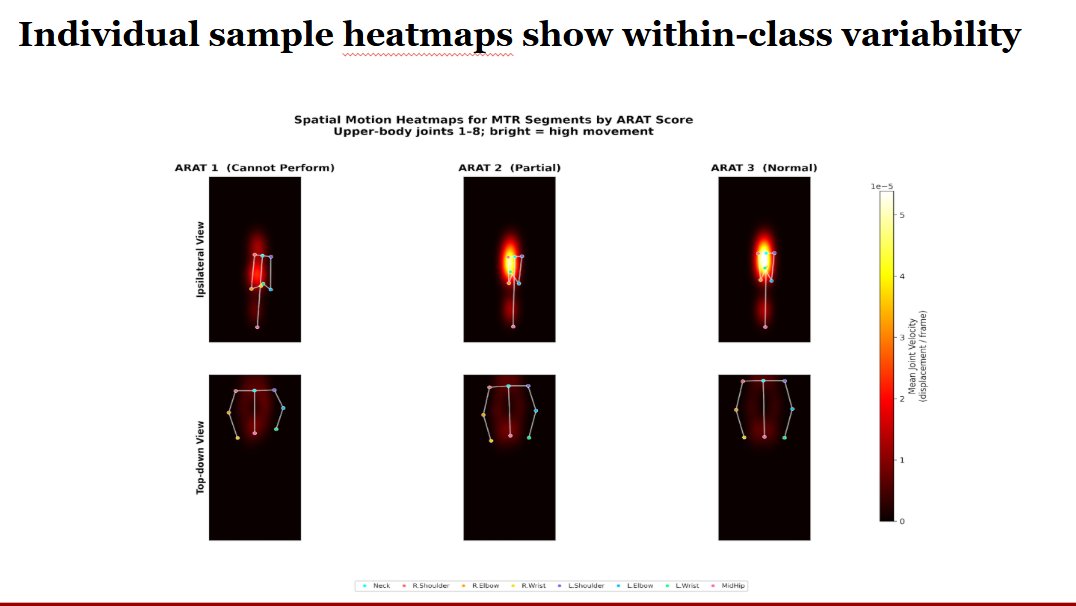}
    \caption{Individual-sample heatmaps (within-class variability).}
  \end{subfigure}
  \caption{\textbf{Extracted motion is interpretable and score-discriminative.}
  (a)~Per-joint motion during manipulation-transport rises with ARAT score and is
  concentrated distally. (b)~Class-mean motion increases and focuses from score 1
  to 3. (c)~Individual exercises vary substantially around the class mean,
  motivating uncertainty-aware composition.}
  \label{fig:outputs}
\end{figure}

Crucially for clinical use, the engine's evidence is inspectable and meaningful (Fig.~\ref{fig:outputs}). During the manipulation-transport phase, mean per-joint velocity rose monotonically with the ARAT score and was largest at the distal joints—wrist and hand—in both views (Fig. \ref{fig:outputs}a), recapitulating the proximal-to-distal organization of upper-limb recovery from features the model extracted without being told to. Class-mean spatial heatmaps showed motion becoming greater and more focused from score 1 to score 3 (Fig.~\ref{fig:outputs}b), while
individual-sample heatmaps revealed substantial within-class variability (Fig.~\ref{fig:outputs}c)—a direct picture of why any single observation is uncertain and why composing many calibrated estimators is the appropriate response rather than a workaround.

\begin{figure}[t]
  \centering
  \includegraphics[width=0.6\textwidth]{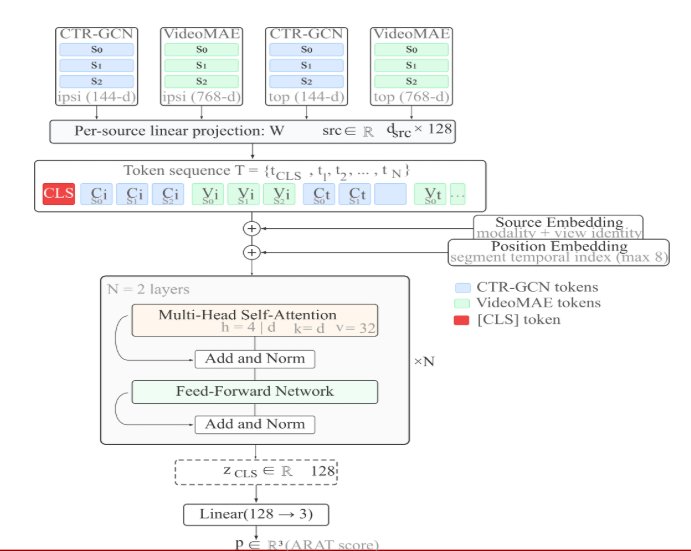}
  \caption{\textbf{Segment-Attention Fusion Transformer (SAFT).} Per-phase tokens
  from CTR-GCN and VideoMAE for the ipsilateral and top-down views are projected to
  a shared 128-dimensional space, tagged with source (modality$+$view) and position
  (phase-index) embeddings, and prepended with a [CLS] token. Two four-head
  self-attention layers attend jointly across modality, view, and phase; the [CLS]
  embedding is mapped by a linear head to the ARAT task score (Methods).}
  \label{fig:saft}
\end{figure}

\subsection*{From a family of expert models to a qualified vocabulary of clinical states}

The engine's output is a qualified clinical state, not a label. For each
patient, the translator assembles a $\mathcal{C}_4$ set spanning all four EAGM layers, and the three gates promote these raw sets to clinically qualified $\mathrm{RC}_4$ states (Table~\ref{tab:rc4}). Across the 105 patients, we observed 667 unique raw $\mathcal{C}_4$ patterns. Of these, 298 passed the compatibility and observability gates to become patient-level $\mathrm{RC}_4$ sets; 164 of those also satisfied cross patient coherence; and 44 reached global $\mathrm{RC}_4$ status—22 high-confidence and 22 low-confidence canonical configurations that recur and generalize across patients. A rating-balanced two-fold cross-validation identified 12 of these as fully robust, present in both folds. The pipeline thus distills tens of thousands of model inferences into a compact, interpretable vocabulary of clinically validated assessment patterns—each a structured explanation spanning environmental context, task score, per-phase scores, and implicated movement-quality elements rather than an opaque number—with 96.1\% of the original predictive uncertainty resolved. Also, for each movement quality element in the $\mathrm{RC}_4$ sets, a weighted kinematic mapping is available on request to support and enhance the decision-making process of the system and the expert together.

\begin{table}[h]
  \centering
  \caption{Qualification of raw compositional assessment sets ($\mathcal{C}_4$) into
  globally validated clinical states ($\mathrm{RC}_4$) through the three-gate pipeline,
  across 88 patients.}
  \label{tab:rc4}
  \begin{tabular}{lc}
    \toprule
    Qualification stage & Sets \\
    \midrule
    Raw $\mathcal{C}_4$ sets                                       & 667 \\
    Patient-level $\mathrm{RC}_4$ (compatibility $+$ observability) & 298 \\
    Cross-patient coherence passed                                 & 164 \\
    Global $\mathrm{RC}_4$ states (22 high-, 22 low-confidence)     & 44  \\
    Robust across two-fold cross-validation                        & 12  \\
    \bottomrule
  \end{tabular}
\end{table}

\subsection*{The family of models reduces the subjectivity of expert scoring}

The cases that most determine whether an automated assessment can be trusted are those where expert scoring is most subjective, and these are common: across doubly-rated data, the two trained raters assigned different scores on 13.7\% of tasks (124 of 904) and 18.1\% of movement-phase segments (615 of 3{,}402; Table~\ref{tab:disagree}). In such a case, there is, by construction, no single ground truth—each of the two scores is endorsed by an expert—so the right target is not to reproduce a non-existent correct answer but to return one a clinician would defend.

The composed family of models does precisely this. In these subjective cases, it returned a score matching at least one of the two raters on 100\% of tasks and 99.2\% of phases (Table~\ref{tab:disagree}), never producing a third, out-of-range value. Measured against either single rater, its accuracy converged near 50\% (Table~\ref{tab:disagree_breakdown})---the signature of a forced choice between two individually valid but conflicting labels, not a failure of the model---and its errors lay almost entirely between adjacent ordinal classes. The engine, therefore, does not impose certainty where none exists; it guaranties a defensible answer within the set that the experts themselves endorse.

How this subjectivity is resolved matters as much as the fact that it exists, and here the composition is doing real work. The calibrated ensemble distributed its agreement between the two raters (for tasks, 0.686 against rater~T1 and 0.315 against~T2), whereas an unweighted product-of-experts collapsed almost entirely onto one rater
(0.016 against~T1 and 0.984 against~T2) while reaching the same match-either rate (Table~\ref{tab:disagree_breakdown}). Both ``resolve'' the subjectivity in the narrow sense, but only the calibrated ensemble does so without silently adopting one expert's idiosyncrasy—reducing subjectivity by landing on a defensible answer
rather than by imposing an arbitrary one.

This behavior is enforced rather than incidental. Before an assessment is reported, a three-gate procedure promotes it from a raw candidate to a clinically qualified one by testing cross-layer compatibility, cross-patient coherence, and evidence observability (Methods); a confidence-based correction that consults the full model family when fusion is uncertain reduced compatibility failures by 66.9\% with no
forced overrides. No candidate failed the observability gate, and an audit of every unique assessment pattern found 79.4\% clinically valid, 9.8\% plausible edge cases, and 10.8\% invalid patterns routed to review. Where adjudication is genuinely ambiguous, the engine returns a score a clinician would defend, flags the residual ambiguity, or defers.

\begin{table}[h]
  \centering
  \caption{Expert subjectivity and its resolution. Two trained raters (T1, T2)
  independently scored each sample; \emph{divergent} cases are those in which their
  scores differed. ``Matches $\geq$1 rater'' is the proportion of divergent cases on
  which the composed ensemble (the Segment-Attention Fusion Transformer for tasks; the
  enhanced-fusion ensemble for phases) reproduced at least one rater's score.}
  \label{tab:disagree}
  \begin{tabular}{lccc}
    \toprule
    Level & Doubly-rated samples & Divergent (\%) & Matches $\geq$1 rater \\
    \midrule
    Task            & 904     & 124 (13.7\%) & 100.0\% \\
    Movement phase  & 3{,}402 & 615 (18.1\%) & 99.2\%  \\
    \bottomrule
  \end{tabular}
\end{table}

\begin{table}[h]
  \centering
  \caption{Agreement with each rater on divergent cases. On cases where the two
  raters diverge, accuracy against either single rater is near 0.50 by construction; the
  calibrated ensemble distributes agreement across both raters, whereas an
  unweighted product-of-experts (PoE) collapses onto one. QWK, quadratic weighted
  Cohen's $\kappa$.}
  \label{tab:disagree_breakdown}
  \footnotesize
  \begin{tabularx}{\textwidth}{@{}lX*{7}{c}@{}}
    \toprule
    & & \multicolumn{3}{c}{Accuracy} & \multicolumn{2}{c}{Weighted $F_1$} & \multicolumn{2}{c}{QWK} \\
    \cmidrule(lr){3-5}\cmidrule(lr){6-7}\cmidrule(lr){8-9}
    Level & Model & T1 & T2 & Avg & T1 & T2 & T1 & T2 \\
    \midrule
    \multirow{2}{*}{Phase ($n=615$)}
      & Enhanced fusion (ensemble) & 0.576 & 0.416 & 0.496 & 0.551 & 0.391 & 0.147 & 0.055 \\
      & Unweighted PoE             & 0.529 & 0.457 & 0.493 & 0.532 & 0.463 & 0.125 & 0.067 \\
    \midrule
    \multirow{2}{*}{Task ($n=124$)}
      & SAFT (ensemble)            & 0.686 & 0.315 & 0.500 & 0.810 & 0.467 & 0.128 & 0.063 \\
      & Unweighted PoE             & 0.016 & 0.984 & 0.500 & 0.031 & 0.986 & 0.045 & 0.742 \\
    \bottomrule
  \end{tabularx}
\end{table}

\subsection*{Clinicians external to development validate the assessments and would adopt them}

Finally, four clinicians with ARAT experience but no role in any phase of
development each independently reviewed 100 automated assessments (400 total), stratified for patient and score diversity, after a one-hour orientation. Review took $2.5\pm0.4$ minutes per task; all four reported a strong fit with clinical practice, and all four indicated a willingness to adopt the system. Per-clinician agreement at every level is given in Table~\ref{tab:validation}. Agreement was unanimous for ARAT scores 0 and 1 and at least 97.5\% for score 2, falling only at score 3 (46.7--81.8\%, mean 67.0\%)—again the near-normal/normal boundary that is the principal locus of both model error and clinician subjectivity. At the phase and movement-quality levels, agreement was high and consistent across phases (80.0--100\% and 80.0--93.9\%, respectively). Clinicians outside the design loop thus concurred strongly with the engine, except at the high-function ceiling where the underlying clinical judgment is itself least certain.

\begin{table}[h]
  \centering
  \caption{Independent clinician validation: agreement (\%) with the automated
  assessments, by level and stratum, for the four clinicians (C1--C4). Task
  agreement is stratified by ARAT score; movement-phase and movement-quality
  agreement are stratified by movement phase (IP, T, MTR, PR). Mean is the
  unweighted mean across the four clinicians.}
  \label{tab:validation}
  \begin{tabular}{llccccc}
    \toprule
    Level & Stratum & C1 & C2 & C3 & C4 & Mean \\
    \midrule
    \multirow{4}{*}{Task (ARAT score)}
      & Score 0 & 100.0 & 100.0 & 100.0 & 100.0 & 100.0 \\
      & Score 1 & 100.0 & 100.0 & 100.0 & 100.0 & 100.0 \\
      & Score 2 & 100.0 & 100.0 & 97.8  & 97.5  & 98.8  \\
      & Score 3 & 81.8  & 66.7  & 46.7  & 72.7  & 67.0  \\
    \midrule
    \multirow{4}{*}{Movement phase}
      & IP  & 98.9 & 97.9 & 86.1 & 100.0 & 95.7 \\
      & T   & 93.0 & 92.6 & 88.6 & 95.0  & 92.3 \\
      & MTR & 90.0 & 85.0 & 80.0 & 92.0  & 86.8 \\
      & PR  & 88.0 & 90.0 & 82.0 & 91.0  & 87.8 \\
    \midrule
    \multirow{4}{*}{Movement quality}
      & IP  & 91.2 & 91.5 & 85.5 & 93.9 & 90.5 \\
      & T   & 90.5 & 82.7 & 84.3 & 89.9 & 86.9 \\
      & MTR & 86.7 & 87.0 & 80.0 & 85.3 & 84.8 \\
      & PR  & 88.0 & 90.0 & 82.0 & 91.0 & 87.8 \\
    \bottomrule
  \end{tabular}
\end{table}

\section*{Discussion}

An engine that composes many calibrated models under probabilistic governance assessed upper-limb function after a stroke at 94.2\% task accuracy ($\kappa=0.934$), reduced predictive uncertainty by 96.1\%, behaved conservatively where expert scoring was subjective, and produced a layered explanation that clinicians outside
the design loop validated and were willing to adopt. The contribution is less a new accuracy number than a demonstration that an automated movement assessment can be made trustworthy and clinically usable—and the path to that runs through three choices that the prevailing approach to medical AI tends to skip.

The first is to treat the ground truth, not the perception model, as the binding constraint. A standard single-clinician ARAT score is a heavily compressed summary of behavior, and fitting a model to it—as the standard collect-labels-then-train pipeline does—bakes in the very information loss that makes the score insensitive to the difference between recovery and compensation~\cite{levin2009motor, van2001responsiveness}. By making the clinician's multi-level judgment expressible, co-design cut labeling entropy by two orders of magnitude and, with the models unchanged, produced an 8.8-point accuracy gain. A phase- and quality-resolved, uncertainty-aware assessment delivered at roughly 2.5 minutes per task is the kind of granular, scalable measurement that could let clinicians titrate therapy and allow trials to detect change that coarse ordinal endpoints miss.

The second is to treat scoring as inference under uncertainty rather than as classification. Recovering a clinical judgment from video is an ill-posed problem over a redundant, high-dimensional behavior, and the appropriate response is not a larger monolithic predictor but the principled composition of many calibrated estimators whose collective entropy can be quantified and acted upon. Simple probabilistic fusion was repeatedly competitive with complex learned fusion, and
only the full composition achieved both high accuracy and near-deterministic confidence; that the extracted features independently recover the proximal-to-distal organization of upper-limb motion (Fig.~\ref{fig:outputs}) indicates the representation is tracking the biomechanics rather than label correlates. The explicit modeling of uncertainty that this enables is precisely what is usually missing from clinical AI~\cite{zhou2025uncertainty}, and it is what lets the system act when confident and defer when it is not.

The third is to design for augmentation rather than replacement. Much of the translational gap in medical AI is attributed not to insufficient accuracy but to systems that are incompatible with clinical reasoning and decision making~\cite{sokol2025translational}. xAARA's decomposition of its output across movement phases and quality elements gives clinicians something to reason with rather than a number to accept, and its calibrated deferral—routing the ambiguous
14–18\% of cases and incoherent patterns to review while matching at least one expert almost everywhere—is the operational form of teaming rather than substitution. Explanation is not automatically beneficial, and poorly matched explanations can induce over-reliance; the value here comes from an explanation pitched at the level of the clinical decision and paired with an honest statement of confidence, which is the configuration under which AI assistance has been shown
to improve rather than distort expert judgment~\cite{zhou2025uncertainty}.

These claims have boundaries. The evaluation is single-session and retrospective; the cohort, though substantial for this domain (88 patients in the integration analysis), is single-context and warrants external, multi-site validation; and the weakest movement-quality elements sit at the frontier of both human and machine
reliability, so their automated detection should be read as decision support, not ground truth. Because xAARA instantiates a domain-agnostic translator, its transfer to other embodied clinical assessments is a hypothesis we are actively testing rather than a result we report. Within those limits, the work offers a concrete template for augmented clinical intelligence: respect the structure and uncertainty of expert practice, make tacit judgment expressible through co-design, and compose many calibrated models under a probabilistic, self-explaining architecture that knows when to defer.

A further consequence of composing under the DBN is that the system yields a structured clinical state rather than a label. The 44 globally qualified $\mathrm{RC}_4$ states are an emergent, data-driven vocabulary of the assessment patterns that recur across stroke survivors, each carrying its own multi-layer explanation and confidence. That this
vocabulary is small enough to be clinically legible, validated to generalize across patients, and structured enough to be reasoned over is what makes it a usable substrate for augmentation: it gives a clinician not a number but a defensible, explainable state—and it is the state space on which a longitudinal tool can be built.

\subsection*{Future directions}
The $\mathrm{RC}_4$ vocabulary turns the static assessment reported here into the state space for an adaptive clinical tool, and two extensions follow directly. The first is \emph{attention recommendation}: each qualified state carries four quality channels—predictive confidence, movement-quality evidence, observability, and inter-rater reliability—and a learnable fusion of these channels can rank which states, movement phases, or quality elements warrant a clinician's focus, with
low-observability or atypical states flagged for verification rather than intervention. Tuned against clinician feedback, this would let the translator not merely report an assessment but direct attention within it. The second is \emph{next-session state prediction}: modeling how patients traverse the $\mathrm{RC}_4$ state space across therapy sessions, conditioned on the recommendation and on time since stroke, would turn assessment into prognosis. This requires prospective, multi-session data and must
respect a fundamental identifiability limit—without recording therapy content, the model cannot separate clinician-directed intervention from spontaneous neurobiological recovery~\cite{krakauer2006motor, borschmann2020recovery}, so future studies should log
exercise content, intensity, and duration to enable that decomposition. Both extensions are formulated here as testable hypotheses; neither is validated by the present single-session data.

\section*{Methods}

\subsection*{Study design and participants}
We studied two cohorts of stroke survivors (Table~\ref{tab:demographics}). A small system-design cohort ($N=2$) took part in the iterative co-design of the capture-and-annotation system. The clinical study then enrolled 105 stroke survivors spanning the full range of upper-limb impairment: ages 23–93 (mean 60.8), 63 male
and 42 female, 50 right- and 55 left-affected, with stroke severity classified as severe ($n=38$), moderate ($n=38$), mild ($n=22$), or recovered ($n=7$). All participants performed the ARAT under therapist supervision. The composition analysis
reported here used the 88-patient subset with complete multi-rater annotation across all assessment levels (788 exercises); the four-clinician validation study used clinicians external to development.


\begin{table}[h]
  \centering
  \caption{Participant demographics for the system-design and clinical-study cohorts.}
  \label{tab:demographics}
  \begin{tabularx}{\textwidth}{@{}lcccX@{}}
    \toprule
    Cohort & Age (years) & Sex & Affected arm & Stroke severity \\
    \midrule
    System design ($N=2$)    & 55--65            & M 2            & L 2            & Severe 1; Mild 1 \\
    Clinical study ($N=105$) & 23--93 ($\mu$ 60.8) & M 63; F 42 & R 50; L 55 & Severe 38; Moderate 38; Mild 22; Recovered 7 \\
    \bottomrule
  \end{tabularx}
\end{table}

\subsection*{Data acquisition}
Each ARAT exercise was recorded with three synchronized cameras providing complementary views: an ipsilateral side view, a top-down view, and a contralateral view zoomed at $2.5\times$ to resolve hand and digit configuration.

\subsection*{Annotation and movement-quality vocabulary}
Each recording was independently scored by two of five participating physical therapists (denoted T1 and T2 in the subjectivity analysis; mean scoring time $2.7\pm0.65$ min per task) using a co-designed hierarchy: a task score and four per-phase scores on $\{0,1,2,3\}$ over the initial-positioning (IP), grasp (T), manipulation-transport (MTR), and placement-release (PR) phases, together with binary ratings for twelve movement-quality composites encoding the qualitative deficits clinicians attend to within each phase. The full co-design and annotation protocol is described separately~\cite{TODO:xAARA}.

\subsection*{Component models}
Three representations were used: CTR-GCN multi-stream skeleton features over body, hand, and object graphs; VideoMAE appearance features per view; and a Hierarchical Bayesian Model pipeline of handcrafted kinematics (joint angles, velocities, smoothness, workspace use, hand aperture). The detector, signal-enhancer (including self-attention imputation), and movement-phase segmentation stages are described in Supplementary Methods. 

\subsection*{Segment-Attention Fusion Transformer (SAFT)}
SAFT (Fig.~\ref{fig:saft}) was the strongest predictor and is, to our knowledge, the novel architectural element of the engine. It takes one token per movement phase from each of four sources (CTR-GCN and VideoMAE for the ipsilateral and top-down views). A per-source linear projection maps each token to a shared 128-dimensional space; the tokens are prepended with a learnable [CLS] token and
given additive source (modality$+$view) and position (phase-index) embeddings. Two transformer encoder layers (four-head self-attention; feed-forward sublayers with residual connections and layer normalization) attend jointly across modality, view, and phase, and the [CLS] embedding ($\in\mathbb{R}^{128}$) is mapped by a linear layer to the task-score logits. By preserving rather than pooling the per-phase structure, SAFT learns to weight the diagnostic phase of a movement, which is why it outperforms pooled task-trained features and conventional fusion classifiers. Training used GroupKFold ($K=5$) by patient with an ordinal focal loss.

\subsection*{The EAGM-DBN translator}
A single end-to-end model cannot deliver the three properties this task demands---a clinically aligned representation, calibrated uncertainty, and a decision-level explanation. The assessment is ill-posed and redundant, the evidence is distributed across modalities, views, and movement phases, and the output must be an interpretable, qualified clinical description rather than a bare score. We therefore compose the component models within a bidirectional Dynamic Bayesian Network (DBN) that operationalizes the EAGM representation as a four-layer probabilistic pipeline---a \emph{translator} between low-level model inferences and the structured assessment a clinician reasons over~\cite{ahmed2025methodology}. In the forward direction, the DBN collects the posteriors of all 692 component models, reconciles their predictions layer by layer through the entropy-gated Product-of-Experts fusion below, and assembles, for each patient--exercise, a compositional assessment set $\mathcal{C}_4 = (e,\, \hat{y}_\text{task},\, (\hat{y}_\text{IP}, \hat{y}_\text{T},
\hat{y}_\text{MTR}, \hat{y}_\text{PR}),\, \{\text{impaired composites}\})$ that carries calibrated confidence at every layer. In the reverse direction, the co-designed structure that defines the EAGM layers acts as a constraint: the three sequential gates below test each $\mathcal{C}_4$ set against the clinicians' cross-layer rules, its coherence across patients, and the observability of its evidence. Critically, the DBN never forces a non-compliant prediction into agreement. Sets it cannot reconcile are flagged to the clinical and development team rather than silently corrected--- compatibility failures are routed to a review list, low-observability sets are marked data-limited, and patterns falling outside the qualified vocabulary are surfaced for expert adjudication. This flagging is the translator's communication channel back to the experts who co-designed the representation, and it is what keeps the system's authority bounded by clinical judgment rather than by the models.

\subsection*{Entropy-gated Product-of-Experts composition}
Component posteriors were combined as
\begin{equation}
  p(\hat{y}=k\mid\mathbf{x}) =
  \frac{\prod_{m} p_m(k\mid\mathbf{x})^{\mathbb{1}[H_{\text{norm}}(p_m)<\tau_H]}}
       {\sum_{k'}\prod_{m} p_m(k'\mid\mathbf{x})^{\mathbb{1}[H_{\text{norm}}(p_m)<\tau_H]}},
\end{equation}
with $H_{\text{norm}}(p_m)=H(p_m)/\log_2 K$ and $\tau_H=0.95$ excluding
near-uniform (uninformative) experts.

\subsection*{Clinical qualification (three gates)}
Each candidate assessment was tested by three sequential gates evaluated over all four EAGM layers: compatibility (consistency with co-designed cross-layer rules), coherence (cross-patient generalization), and observability (sufficient evidence at every layer). When fusion confidence was low ($\max\mathbf{p}<0.5$), a correction step consulted the full model family; incompatible candidates were routed to review rather than forced into compliance.

\subsection*{Uncertainty, agreement, and subjectivity analysis}
Labeling uncertainty used a hierarchical Bayesian label model (reported in nats). Predictive uncertainty used normalized entropy over 3-fold cross-validation. Agreement is Cohen's $\kappa$ (quadratic weighting, QWK, where noted). Divergent cases—those on which the two raters T1 and T2 assigned different scores—were held out of training
and scored against each rater separately, against their average, and by whether the prediction matched either rater.

\subsection*{Statistical analysis}
Accuracies are reported with Cohen's $\kappa$ (task and phase levels) and weighted $F_1$ (movement-quality composites); ordinal agreement on divergent cases uses quadratic weighted $\kappa$. Cross-validation was grouped by patient (GroupKFold,
$K=5$) throughout to prevent leakage, and predictive-uncertainty statistics are means $\pm$ s.d. over 3-fold cross-validation. Sample sizes ($n$) denote analyzed exercises, phase segments, or composites as indicated in each table.

\subsection*{Validation study}
Four clinicians not involved in development each reviewed 100 stratified automated assessments after a one-hour orientation; we recorded review time, qualitative fit,
and willingness to adopt, and computed per-clinician agreement at the task, phase, and movement-quality levels.

\section*{Data availability}
\section*{Code availability}

 \bibliography{sn-bibliography}   

\end{document}